\documentclass[sigconf,natbib=true]{acmart}
\AtBeginDocument{%
  }

\usepackage{arydshln}
\usepackage{lipsum}
\usepackage{tabularray}
\usepackage{multirow}
\usepackage{booktabs, caption, array}
\usepackage{algorithm}
\usepackage{algorithmic}
\usepackage{comment}
\usepackage{xcolor}
\usepackage{subcaption}
\usepackage{graphicx}
\usepackage{subcaption}
\usepackage{enumitem}
\usepackage{framed}
\usepackage{hyperref} 
\usepackage{amsmath}
\definecolor{shadecolor}{gray}{0.9}
\usepackage{comment}
\usepackage{xspace}
\usepackage{colortbl}
\usepackage{xcolor}
\usepackage{graphicx}
\usepackage{subcaption}  
\definecolor{o_blue}{HTML}{FFF4E6}
\newcommand{\thl}[2][o_blue]{%
  \colorbox{#1}{\strut\hspace{1pt}#2\hspace{1pt}}%
}
\copyrightyear{2026}
\acmYear{2026}
\setcopyright{cc}
\acmISBN{978-1-4503-XXXX-X/2018/06}

\begin{document}

\title{Deep Search with Hierarchical Meta-Cognitive Monitoring Inspired by Cognitive Neuroscience}

\author{Zhongxiang Sun}
\author{Qipeng Wang}
\affiliation{
  \institution{\mbox{Gaoling School of Artificial
Intelligence}\\Renmin University of China}
  \city{Beijing}\country{China}
  }
\email{{sunzhongxiang, wqp}@ruc.edu.cn}

\author{Weijie Yu}
\affiliation{
  \institution{\mbox{School of Information Technology} \\ and Management\\University of International Business and Economics
}
  \city{Beijing}\country{China}
  }
\email{yu@uibe.edu.cn}

\author{Jingxuan Yang}
\affiliation{%
  \institution{Search Applications Department, Tencent}
  \city{Beijing}\country{China}
  }
\email{emmajxyang@tencent.com}

\author{Haolang Lu}
\affiliation{%
  \institution{Beijing University of Posts and Telecommunications}
  \city{Beijing}\country{China}
  }
\email{lhl_bupt@bupt.edu.cn}

\author{Jun Xu}
\affiliation{%
  \institution{\mbox{Gaoling School of Artificial
Intelligence}\\Renmin University of China}
  \city{Beijing}\country{China}
  }
\email{junxu@ruc.edu.cn}

\renewcommand{\shortauthors}{Zhongxiang Sun et al.}

\begin{abstract}
Deep search agents powered by large language models have demonstrated strong capabilities in multi-step retrieval, reasoning, and long-horizon task execution. However, their practical failures often stem from the lack of mechanisms to monitor and regulate reasoning and retrieval states as tasks evolve. Insights from cognitive neuroscience suggest that human metacognition is hierarchically organized, integrating fast anomaly detection with selectively triggered, experience-driven reflection.
In this work, we propose \textbf{Deep Search with Meta-Cognitive Monitoring (DS-MCM)}, a deep search framework augmented with an explicit hierarchical metacognitive monitoring mechanism. DS-MCM integrates a \textbf{Fast Consistency Monitor}, which performs lightweight checks on the alignment between external evidence and internal reasoning confidence, and a \textbf{Slow Experience-Driven Monitor}, which is selectively activated to guide corrective intervention based on experience memory from historical agent trajectories. By embedding monitoring directly into the reasoning–retrieval loop, DS-MCM determines both when intervention is warranted and how corrective actions should be informed by prior experience.
Experiments across multiple deep search benchmarks and backbone models demonstrate that DS-MCM consistently improves performance and robustness. 
\end{abstract}


\maketitle

\section{Introduction}
\begin{figure}[t] 
  \centering
       \includegraphics[width=0.99\linewidth]{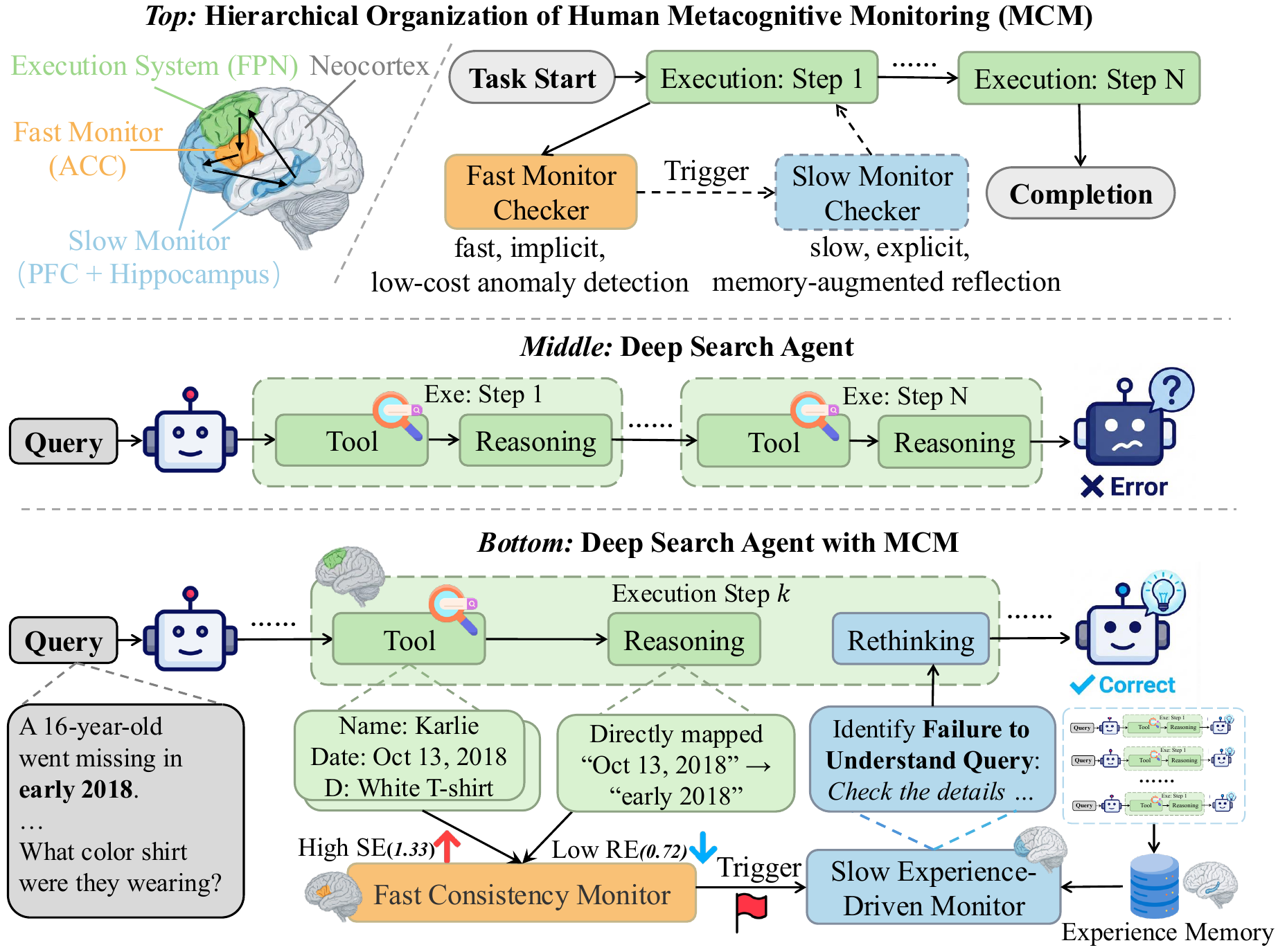} 
  \caption{Cognitive neuroscience-inspired hierarchical metacognitive monitoring for deep search agents. \textit{Top:} Humans rely on a fast, implicit monitor to detect anomalies during execution, which selectively triggers a slow, experience-driven monitor for reflective correction. \textit{Middle:} Standard deep search agents lack such monitoring, allowing early cognitive errors to propagate across reasoning steps. \textit{Bottom:} Our DS-MCM integrates the fast consistency monitor and slow experience-driven monitor into the deep search loop, enabling timely detection and correction of unreliable execution (Illustrative examples sampled from BrowseComp-Plus~\cite{chenbrowsecomp}).}
  \label{fig:introduction}
\end{figure}

Human problem solving relies not only on strong reasoning ability, but also on the capacity to \textbf{monitor and regulate one’s own reasoning process}~\cite{nelson1990metamemory}. Through evolution, humans have developed cognitive systems that can continuously detect when reasoning becomes unreliable, adjust confidence in response to ambiguous evidence, and correct errors before they propagate~\cite{metcalfe2013evolution,fleming2014measure}. This \textbf{metacognitive capability} enables robust decision making in complex and uncertain environments~\cite{gehring1993neural}.

In contrast, modern deep search agents powered by large language models have achieved remarkable progress in retrieval, reasoning, and long-horizon planning~\cite{yao2022react,team2025tongyi}. By iteratively interacting with external tools and synthesizing information across multiple steps, deep search agents can solve increasingly complex, open-ended tasks. However, despite their strong cognitive capacity, these agents still exhibit systematic failures in practice. These failures tend to arise not at the level of isolated reasoning steps, but as tasks evolve with intermediate feedback and partial or conflicting evidence~\cite{zhang2025far}. In practice, agents are often seen to follow rigid reasoning trajectories with limited adjustment to newly acquired information, while retrieval behaviors may treat information acquisition, integration, and verification as loosely connected stages~\cite{zhang2025far,shi2025deep}. Such patterns suggest a broader phenomenon in which agents struggle to \emph{maintain and regulate} their reasoning and retrieval states over time, especially under uncertainty, rather than a simple lack of reasoning or retrieval capacity.

Insights from cognitive neuroscience suggest that human metacognition is organized hierarchically~\cite{metcalfe2013evolution}. As shown in the top of~\autoref{fig:introduction}, rather than relying on a single reflective process, humans employ a \textbf{fast monitoring mechanism} that rapidly detects anomalies such as conflict or prediction error~\cite{gehring1993neural,botvinick2001conflict}, and a \textbf{slow monitoring mechanism} that performs deliberate reflection and corrective control when needed~\cite{shea2014supra}. Crucially, the slow mechanism is not invoked continuously; it is selectively triggered by signals from the fast monitor and is shaped by accumulated experience, rather than operating as a generic, context-free evaluator~\cite{shenhav2013expected,flavell1979metacognition}.

Existing approaches to agent monitoring provide partial signals for detecting abnormal behavior, but remain insufficient for the demands of {deep search agents}, where reasoning and retrieval are tightly interleaved over multiple steps. \textbf{Fast monitoring} is commonly approximated using token-level uncertainty measures~\cite{malinin2020uncertainty,renout}, which can be unreliable in deep search settings where \textit{multiple reasoning paths from diverse retrieval results may be simultaneously plausible}, such that high entropy does not necessarily correspond to erroneous behavior~\cite{dong2025agentic}. Conversely, low entropy may instead signal overconfident reasoning that fails to consider alternative retrieved evidence. \textbf{Slow monitoring} is typically implemented through a standalone critical model that evaluates reasoning in isolation, \textit{without access to historical experience or accumulated knowledge about recurring failure and successful patterns}~\cite{goucritic,liu2025inference}, in contrast to human reflective processes that are shaped by experience~\cite{metcalfe2013evolution,nelson1990metamemory}.


In this work, we propose \textbf{Deep Search with Meta-Cognitive Monitoring (DS-MCM)}, a deep search framework augmented with an explicit hierarchical metacognitive monitoring mechanism. As shown in \autoref{fig:introduction}, our approach decomposes metacognitive monitoring into a \textbf{Fast Consistency Monitor} and a \textbf{Slow Experience-Driven Monitor}, tightly integrated into the agent’s reasoning--retrieval loop. \textit{(1) The fast monitor} tracks the alignment between external \textbf{Searching Entropy} \textbf{(SE)} and internal \textbf{Reasoning Entropy} \textbf{(RE)}, enabling the detection of abnormal execution states where confidence and evidence diverge. \textit{(2) The slow monitor} draws on metacognitive patterns distilled from historical agent trajectories to supervise and regulate deepsearch execution at test time. Together, this design allows monitoring signals to go beyond raw uncertainty estimates and context-agnostic critique, providing principled guidance on \textbf{when intervention is warranted} and \textbf{how corrective actions should be informed by past experience}.
Experiments across multiple deep-search benchmarks and agent architectures demonstrate consistent performance improvements.

The contributions of this paper are summarized as follows:

(1) We introduce a \emph{neuroscience-inspired perspective} on monitoring deep search agents, highlighting that existing systems lack explicit metacognitive mechanisms to maintain and regulate their reasoning and
retrieval states. This perspective motivates the need for structured monitoring beyond task-level reasoning and retrieval.

(2) We propose \textbf{Deep Search with Meta-Cognitive Monitoring (DS-MCM)}, a hierarchical framework for deep-search agents that separates task-level reasoning from metacognitive control by combining fast, evidence-aware consistency monitoring with slow, experience-driven reflective regulation for deep-search agent.

(3) We empirically demonstrate the effectiveness of metacognitive monitoring in deep search, showing consistent improvements in robustness and reasoning resilience across multiple benchmarks and backbone models, and enabling open-source systems to match or surpass strong proprietary deep-search systems.

\section{Related Work}
\subsection{Deep Search Agents}
Recent advances in large language models have enabled \emph{Deep Search} (or \emph{Deep Research}) agents, which address complex, long-horizon information-seeking tasks through iterative interaction with external sources. Unlike single-shot question answering, these agents operate in open-world settings and tightly interleave reasoning with web search, browsing, and evidence synthesis~\cite{shi2025deep,team2025tongyi}.
The development of Deep Search agents has been closely coupled with benchmarks emphasizing persistent web interaction and multi-step information gathering, such as GAIA and BrowseComp~\cite{mialon2023gaia,wei2025browsecomp}. These benchmarks have motivated agent architectures that integrate large reasoning models with explicit search and browsing modules, typically organized as structured reasoning--action loops and often optimized via reinforcement learning or preference-based objectives~\cite{yao2022react,li2025webthinker}. Recent studies further demonstrate that increasing the depth of agent--environment interaction can substantially improve performance, even without scaling model size or context length~\cite{team2025mirothinker}.
Despite this progress, most existing Deep Search agents primarily optimize task-level outcomes and lack explicit mechanisms for monitoring or regulating their own reasoning processes. They often assume that deeper retrieval or longer reasoning chains will resolve uncertainty, whereas empirical analyses reveal frequent unstable behaviors during execution, including over-confident reasoning under ambiguous evidence and inefficient or repetitive search patterns~\cite{zhang2025far}. In contrast, our work introduces \textbf{metacognitive monitoring} as a core capability for Deep Search agents, enabling the detection of abnormal execution states and the regulation of search behavior during interaction.

\subsection{Critical Models}
\label{sec:critical_models}

Recent work has explored \emph{process-level supervision and critique mechanisms} for large language models to detect and correct reasoning errors during execution. A representative direction is \textbf{Process Reward Models (PRMs)}, which provide step-wise supervision signals for intermediate reasoning states, but suffer from limitations such as annotation noise, reward bias, and weak cross-task generalization~\cite{zhang2025lessons}. Benchmarks such as ProcessBench further formalize step-level error localization~\cite{zheng2025processbench}, while PRM-style supervision has been extended to retrieval-augmented and multi-step reasoning settings~\cite{sun2025rearter}.  
In parallel, \textbf{LLM-as-critic} approaches generate structured critique via separate critic models or self-critique mechanisms, improving robustness without dense human supervision~\cite{xi2024enhancing,yu2025self}. Orthogonal to explicit critics, \textbf{uncertainty- and entropy-based detection methods} provide lightweight signals for identifying unreliable reasoning or hallucinations. These approaches leverage measures such as semantic entropy or representation-level uncertainty probes to flag potentially incorrect generations without requiring step-level supervision~\cite{malinin2020uncertainty,farquhar2024detecting}. Recent studies further extend critical and uncertainty-aware mechanisms to long-horizon and agentic reasoning, showing that such signals can be incorporated into extended reasoning trajectories, though they are often reactive and local in scope~\cite{sun2025detection}.

Despite their effectiveness at step-level oversight, these methods are largely designed for standalone reasoning traces or generic correction loops. They lack explicit \textbf{metacognitive monitoring mechanisms} for deep search agents operating under evolving, incomplete, and potentially inconsistent external evidence. In particular, prior work does not model the {calibration between internal reasoning confidence and external evidence uncertainty}, nor does it leverage \textbf{historical execution experience} to guide monitoring and regulation.  This limitation prevents agents from dynamically identifying abnormal search states and adjusting their behavior in a human-like manner during long-horizon execution. 
In contrast, our work introduces a \textbf{hierarchical metacognitive monitoring framework} tailored to deep search agents, combining fast uncertainty-consistency checks with slow, experience-driven reflection. This design enables lightweight online detection of abnormal reasoning states and sustained, agent-level regulation over long-horizon deep search processes.


\section{Methods}
\begin{figure*}[t] 
  \centering
  \includegraphics[width=0.99\linewidth]{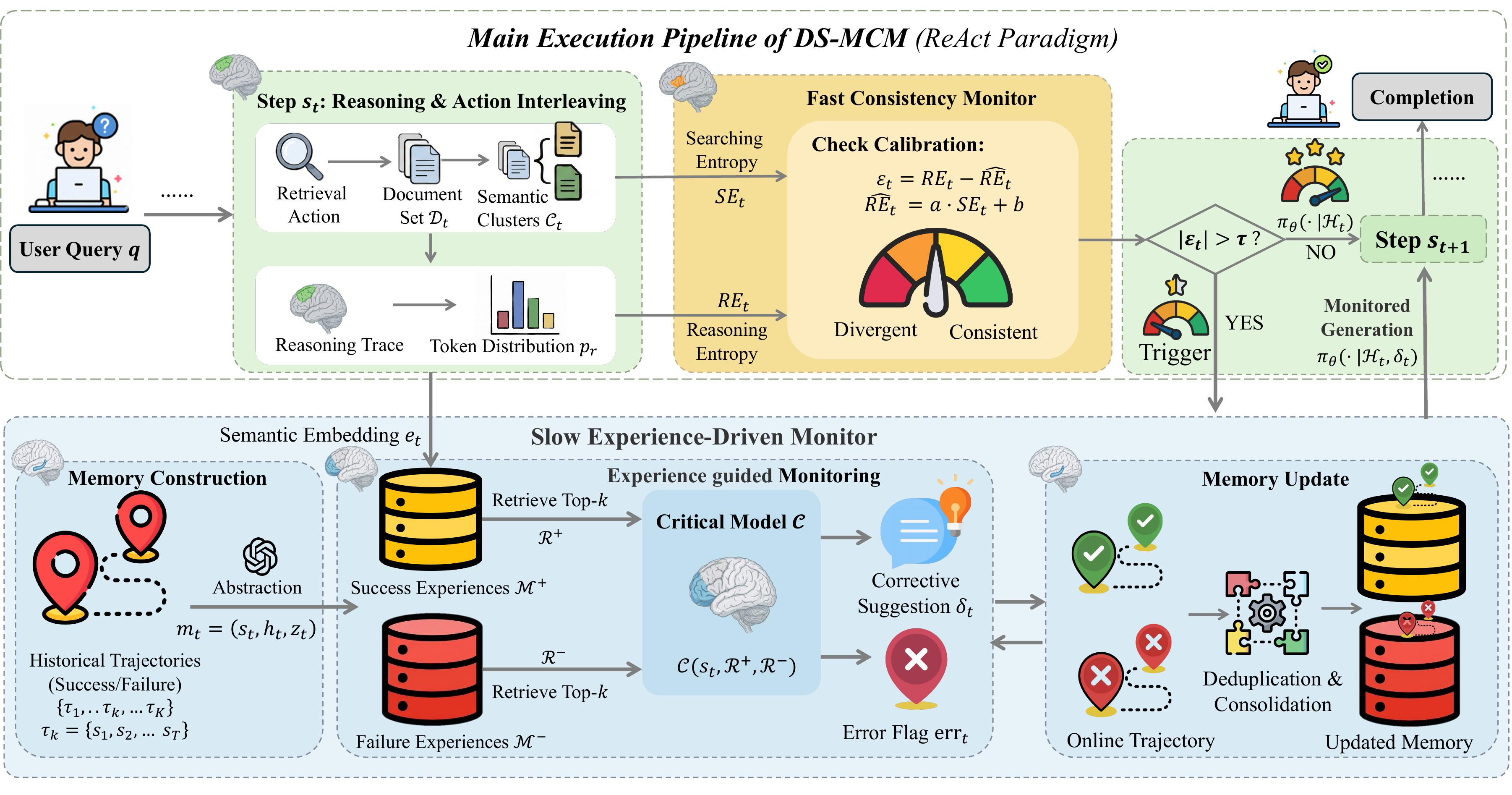} 
  \caption{Overview of Deep Search with Meta-Cognitive Monitoring (DS-MCM).
DS-MCM augments a standard ReAct-based deep search agent with an explicit metacognitive monitoring module. 
At each step, a Fast Consistency Monitor~(\S\ref{sec:fast_monitor}) checks whether reasoning uncertainty is calibrated with the uncertainty of retrieved evidence. When abnormal mismatches are detected, a Slow Experience-Driven Monitor~~(\S\ref{sec:slow_monitor}) is triggered to perform reflective diagnosis using success and failure experiences stored in metacognitive memory, and to issue corrective guidance for subsequent steps. The memory is continuously updated and consolidated online, enabling efficient and experience-driven regulation of deep search execution.}
  \label{fig:method}
\end{figure*}
\subsection{Overview}
\label{sec:overview}

We consider a \textbf{deep search agent} $\pi_{\theta}$ following the ReAct paradigm~\cite{yao2022react}, where reasoning and information acquisition are interleaved through iterative interactions with external tools. Given a user query $q$, the agent executes a sequence of steps
\begin{equation}
\mathcal{H}_T^{q} = \{ s_1, s_2, \dots, s_T \},
\end{equation}
where each step $s_t$ consists of (i) internal reasoning generated by the base language model, and (ii) an action such as a tool call or a termination decision. Tool outputs are incorporated into the context and inform subsequent reasoning steps.
Despite strong reasoning and retrieval capabilities, standard deep search agents lack an explicit mechanism to \textbf{monitor and regulate their own execution state} as the search process unfolds. To address this limitation, we propose \textbf{Deep Search with Meta-Cognitive Monitoring (DS-MCM)}, which augments the standard deep search loop with an explicit monitoring layer. As illustrated in Figure~\ref{fig:method}, {the key design principle of DS-MCM is to separate \textbf{task-level cognition} (reasoning and retrieval) from \textbf{metacognitive control}, while tightly integrating the latter into the execution process.}
DS-MCM introduces two complementary monitoring components. A \textbf{Fast Consistency Monitor} operates at every step with low overhead, providing rapid signals about potential inconsistencies between internal reasoning uncertainty and external evidence uncertainty. When such signals indicate abnormal execution states, a \textbf{Slow Experience-Driven Monitor} is selectively triggered to perform reflective analysis and generate corrective guidance based on accumulated metacognitive experience.
At each step, monitoring is applied after internal reasoning and before the next action decision, allowing the agent to either proceed normally or adjust its behavior when necessary. This hierarchical design enables continuous yet non-intrusive regulation of deep search execution, ensuring that reflective intervention is invoked only when warranted.

\subsection{Fast Consistency Monitor}
\label{sec:fast_monitor}

Unlike single-shot generation, deep search agents iteratively interleave reasoning with evidence acquisition, causing the agent’s internal reasoning state to be continuously shaped by retrieved information. As a result, uncertainty observed during reasoning is not inherently pathological. When external evidence is ambiguous, incomplete, or mutually inconsistent, it is natural and often desirable for the agent to maintain multiple competing hypotheses. In such cases, elevated reasoning uncertainty reflects legitimate multi-path exploration rather than failure~\cite{dong2025agentic}.

This observation implies that reasoning uncertainty should not be evaluated in isolation. Instead, it must be interpreted \textbf{relative to the uncertainty of the retrieved evidence}. An agent may reasonably exhibit high uncertainty when the evidence itself is unclear, while persistent uncertainty under clear and consistent evidence often signals abnormal execution states, such as information misuse, tool misalignment, or unstable reasoning dynamics. Conversely, overly confident reasoning in the presence of ambiguous evidence may indicate premature commitment and ignored counter-evidence.

These considerations motivate a \textbf{consistency-based monitoring principle}: a deep search agent operates normally when its internal reasoning uncertainty is calibrated to the uncertainty of external evidence, and becomes suspicious when the two diverge. This perspective departs from conventional fast monitoring approaches that treat token-level entropy as a direct proxy for correctness. In deep search settings, high token entropy may simply reflect evidence-induced ambiguity, while low entropy may still correspond to confident but incorrect conclusions. \emph{The Fast Consistency Monitor is therefore explicitly designed for deep search execution, modeling the expected relationship between retrieval uncertainty and reasoning uncertainty rather than relying on raw token statistics alone.}

\subsubsection{Searching Entropy (SE)}
Formally, let $q$ denote the user query and let $t$ index a ReAct step $s_{t}$. After executing a retrieval action at step $t$, the agent obtains a set of retrieved documents
\begin{equation}
\mathcal{D}_t = \{ d_1, \dots, d_K \}.
\end{equation}

Rather than treating individual documents as independent evidence, we characterize the \textbf{semantic diversity} of the retrieved results by operating in an embedding space. Specifically, each document $d_k$ is mapped to a dense semantic representation $\mathbf{e}_k$ using a pretrained embedding model~\cite{qwen3embedding}. Documents that convey similar semantic content are expected to form coherent neighborhoods in this space.

To approximate the distribution of {semantic interpretations implied by the retrieved evidence}, we induce a embedding based clustering over the documents based on their embedding similarity~\cite{Campello2013DensityBasedCB}. This yields a set of latent semantic clusters
\begin{equation}
\mathcal{C}_t = \{ c_1, \dots, c_M \},
\end{equation}
where each cluster $c_i$ represents a distinct semantic theme or evidence pattern present in the retrieved results.

We define a probability distribution over these semantic clusters by aggregating the normalized contribution of documents assigned to each cluster,
\begin{equation}
p_s(c_i \mid q, \mathcal{D}_t),
\qquad
\sum_{i=1}^{M} p_s(c_i \mid q, \mathcal{D}_t) = 1,
\end{equation}
where $p_s(c_i \mid q, \mathcal{D}_t)$ reflects the relative semantic mass of cluster $c_i$ in the retrieved evidence.

We then define the \textbf{Searching Entropy (SE)} at step $t$ as the Shannon entropy of this semantic distribution:
\begin{equation}
\mathrm{SE}_t
=
-\sum_{i=1}^{M}
p_s(c_i \mid q, \mathcal{D}_t)
\log p_s(c_i \mid q, \mathcal{D}_t).
\end{equation}

Intuitively, a low value of $\mathrm{SE}_t$ indicates that the retrieved documents are semantically coherent and concentrate around a small number of consistent evidence patterns, whereas a high value reflects \textbf{semantic fragmentation}, where the retrieval results support multiple competing or incompatible interpretations.

\subsubsection{Reasoning Entropy (RE)}
To capture the agent’s internal uncertainty during reasoning, we define \textbf{Reasoning Entropy (RE)} based on the base language model’s token prediction distributions within the reasoning segment. Let the reasoning trace at step $t$ consist of $T$ generation positions. At each position $j$, we consider the normalized next-token distribution restricted to the top-$K$ tokens, denoted by $p_r(\cdot)$. The Reasoning Entropy is computed as
\begin{equation*}
\mathrm{RE}_t
=
\frac{1}{T}
\sum_{j=1}^{T}
\left(
-\sum_{k=1}^{K}
p_r\!\left(
y_{t,j}^{(k)} \mid y_{<j}, q, \mathcal{D}_t
\right)
\log
p_r\!\left(
y_{t,j}^{(k)} \mid y_{<j}, q, \mathcal{D}_t
\right)
\right).
\end{equation*}

Low values of $\mathrm{RE}_t$ indicate stable and confident reasoning trajectories, whereas high values suggest uncertainty among multiple plausible reasoning continuations.

\subsubsection{Fast Consistency Monitor.}
The Fast Consistency Monitor detects abnormal execution states by assessing whether the observed reasoning uncertainty $\mathrm{RE}_t$ is consistent with the uncertainty of the retrieved evidence $\mathrm{SE}_t$. Using a set of historically successful steps from the same memory set of the Slow Experience-Driven Monitor (Section~\ref{sec:slow_monitor}), we fit a simple calibration function that captures the typical relationship between evidence uncertainty and reasoning uncertainty,
\begin{equation}
\widehat{\mathrm{RE}}_t = a \cdot \mathrm{SE}_t + b,
\end{equation}
where $a\ge 0$ and $b$ are learnable parameters. 
We then define the residual:
\begin{equation}
\epsilon_t = \mathrm{RE}_t - \widehat{\mathrm{RE}}_t,
\end{equation}
where the residual $\epsilon_t$ quantifies the degree of mismatch between internal and external uncertainty. Small residuals indicate well-calibrated reasoning behavior, whereas large positive residuals correspond to unusually high reasoning uncertainty given clear evidence, and large negative residuals reflect unusually confident reasoning despite ambiguous evidence.

In practice, we estimate the standard deviation $\sigma$ of residuals from successful trajectories and define an anomaly threshold $\tau = k\sigma$. A step is flagged as anomalous when
\begin{equation}
|\epsilon_t| > \tau.
\end{equation}
When such an anomaly is detected, the Fast Consistency Monitor triggers the Slow Experience-Driven Monitor, which performs reflective diagnosis and issues corrective control signals (Section~\ref{sec:slow_monitor}).

Overall, the Fast Consistency Monitor provides a lightweight, step-level mechanism that grounds internal reasoning uncertainty in the uncertainty of retrieved evidence, making it particularly well suited to the iterative retrieval--reasoning dynamics of deep search agents.

\subsection{Slow Experience-Driven Monitor}
\label{sec:slow_monitor}

The Fast Consistency Monitor provides a lightweight mechanism for detecting \emph{when} a deep search step deviates from normal uncertainty calibration. However, identifying \emph{why} such deviations arise and \emph{how} the agent should adapt its behavior requires reflective reasoning grounded in prior experience. To address this need, we introduce a \textbf{Slow Experience-Driven Monitor}, which supervises deep search execution using a persistent metacognitive memory distilled from historical agent trajectories.

Formally, the Slow Monitor operates over a memory set \( \mathcal{M} \) constructed from past executions and interacts with the current trajectory
\(
\mathcal{H}_t = \{ s_1, \dots, s_t \}
\)
in a selective, memory-conditioned manner. Unlike the Fast Monitor, which relies solely on local uncertainty signals, the Slow Monitor performs experience-informed diagnosis and generates corrective guidance by comparing the current execution state against previously observed cognitive patterns.

\subsubsection{Metacognitive Experience Memory Construction}

The Slow Experience-Driven Monitor is supported by a compact metacognitive memory that encodes \textbf{reusable cognitive experience} extracted from historical deep search executions. Rather than storing raw trajectories or long-horizon interaction histories, we distill past executions into lightweight, session-level memory entries that emphasize \emph{how} the agent reasoned and interacted with tools at each step.

We construct this memory offline from historical trajectories. Given a deep search execution
\begin{equation}
\tau_k = \{ s_1, s_2, \dots, s_T \},
\end{equation}
each step \( s_t \) corresponds to one ReAct iteration and is treated as a \textbf{session-level cognitive unit}. A session preserves only local information relevant for metacognitive analysis, including the user query, the agent’s internal reasoning trace, executed action, and tool feedback. This representation deliberately abstracts away task-specific surface details while retaining the agent’s cognitive behavior at that step.

Supervision for historical executions is typically available only at the trajectory level. To enable scalable memory construction, we propagate trajectory-level outcomes to all constituent sessions:
\begin{equation}
y(s_t) =
\begin{cases}
1, & \text{if the trajectory is successful}, \\
0, & \text{if the trajectory is unsuccessful}.
\end{cases}
\end{equation}
This coarse-grained labeling reflects our focus on \textbf{global cognitive behavior patterns}. In successful trajectories, intermediate steps usually exhibit coherent reasoning and tool usage that collectively support task completion. In unsuccessful trajectories, cognitive deficiencies tend to recur across steps, even when isolated actions appear locally reasonable.

Each session is then converted into a metacognitive memory entry through a label-conditioned abstraction process. For a session \( s_t \), we generate
\begin{equation}
m_t = \big( s_t,\; h_t,\; z_t,\; y(s_t) \big),
\end{equation}
where \( s_t \) denotes the structured representation of the current session, \( h_t \) is a concise summary of historically relevant sessions leading up to \( s_t \), and \( z_t \) is a natural-language abstraction extracted by LLMs capturing the cognitive behavior expressed at step \( t \). 

The abstraction of \( z_t \) is explicitly \textbf{label-dependent}. For sessions labeled as successful (\(y(s_t)=1\)), the abstraction prompt emphasizes the extraction of \textbf{positive cognitive behaviors}, guiding the model to identify effective reasoning strategies, appropriate tool usage, and sound evidence integration patterns. For sessions labeled as unsuccessful (\(y(s_t)=0\)), the abstraction focuses on \textbf{failure diagnosis}, encouraging the model to surface recurring cognitive deficiencies, such as insufficient external information acquisition, premature closure, or verification mechanism failure, and to articulate high-level improvement insights~\cite{zhang2025far}.

Although the two abstraction prompts share an identical output structure, their semantic emphasis induces systematically different representations. As a result, all memory entries are organized into two complementary repositories:
\begin{equation}
\mathcal{M} = \mathcal{M}^{+} \cup \mathcal{M}^{-},
\end{equation}
where \( \mathcal{M}^{+} \) stores \textbf{success experiences} that exemplify effective cognitive behaviors, and \( \mathcal{M}^{-} \) stores \textbf{failure experiences} that capture typical error patterns together with their corrective insights. These two memory spaces provide the experiential foundation upon which the Slow Experience-Driven Monitor performs memory-conditioned diagnosis and intervention.

\subsubsection{Memory-Conditioned Monitoring and Intervention}

When the Fast Consistency Monitor flags a session \( s_t \) as potentially abnormal, the Slow Experience-Driven Monitor is selectively activated to interpret the anomaly through prior experience. At this stage, the current session is used as a query to retrieve relevant experiences from both success and failure memory spaces.

We encode the current session representation into a dense semantic embedding
\begin{equation}
\mathbf{e}_t = f_{\text{enc}}(s_t),
\end{equation}
where \( f_{\text{enc}}(\cdot) \) denotes a fixed embedding model~\cite{qwen3embedding}. For each memory entry \( m_i \in \mathcal{M} \), we precompute an embedding
\begin{equation}
\mathbf{e}_i(m_i) = f_{\text{enc}}(s_i)+f_{\text{enc}}(h_i),
\end{equation}
where \( h_i \) is the stored historical-session summary associated with that memory entry.

Retrieval is performed independently over the success and failure memory pools by computing cosine similarity:
\begin{equation}
\mathrm{sim}(s_t, m_i)
=
\cos\!\left(\mathbf{e}_t,\; \mathbf{e}_i\right),
\qquad
m_i \in \mathcal{M}^{+} \text{ or } \mathcal{M}^{-}.
\end{equation}

For each memory pool, we select the top-\(K\) most similar entries:
\begin{equation}
\mathcal{R}_t^{+}
=
\operatorname*{TopK}_{m_i \in \mathcal{M}^{+}}
\mathrm{sim}(s_t, m_i),
\qquad
\mathcal{R}_t^{-}
=
\operatorname*{TopK}_{m_i \in \mathcal{M}^{-}}
\mathrm{sim}(s_t, m_i).
\end{equation}

Conditioned on the current session and the retrieved success and failure experiences, the \textbf{Critical Model} \( \mathcal{C} \) performs reflective evaluation. Formally, it defines a mapping
\begin{equation}
(\mathrm{err}_t,\; \delta_t)
=
\mathcal{C}\!\left(
s_t,\; \mathcal{R}_t^{+},\; \mathcal{R}_t^{-}
\right),
\end{equation}
where \( \mathrm{err}_t \in \{0,1\} \) indicates whether a cognitive error is identified at step \( t \), and \( \delta_t \) denotes a corresponding corrective suggestion. When \( \mathrm{err}_t = 0 \), we set \( \delta_t = \varnothing \).

The corrective suggestion \( \delta_t \) modulates subsequent deep search execution by conditioning the base agent’s policy:
\begin{equation}
s_{t+1}
\sim
\pi_{\theta}\!\left(
\cdot \mid \mathcal{H}_{t},\; \delta_t
\right),
\end{equation}
where \( \mathcal{H}_{t} = \{s_1,\ldots,s_t\} \) denotes the updated execution context. When no cognitive error is detected, policy execution reduces to standard deep search behavior without intervention.

\subsubsection{Online Memory Update and Consolidation (optional)}

During online execution, ground-truth trajectory-level correctness labels are unavailable. Instead, metacognitive memory is updated incrementally based on signals provided by the Fast Consistency Monitor and the Slow Experience-Driven Monitor.

After completing a session \( s_t \), the resulting online supervision signal is denoted as
\begin{equation}
y_t^{\text{online}} \in \{\text{success},\; \text{failure}\}.
\end{equation}

For each online-labeled session, we construct a metacognitive memory entry
\begin{equation}
m_t = \big( s_t,\; h_t,\; z_t \big),
\end{equation}
which is assigned to either the success memory \( \mathcal{M}^{+} \) or the failure memory \( \mathcal{M}^{-} \).

To prevent uncontrolled memory growth, we perform online deduplication by computing embedding similarity:
\begin{equation}
\mathrm{sim}(m_t, m_i)
=
\cos\!\big(\mathbf{e}(m_t),\; \mathbf{e}(m_i)\big).
\end{equation}
If
\begin{equation}
\max_{m_i} \mathrm{sim}(m_t, m_i) \ge \tau_{\text{dup}},
\end{equation}
the candidate is discarded; otherwise, it is inserted into the corresponding memory pool.

\section{Experiments}

In this section, we evaluate DS-MCM by answering the following research questions:

\noindent\textbf{RQ1: Overall Effectiveness.}
Does DS-MCM improve the performance of deep-search agents across benchmarks and backbones?

\noindent\textbf{RQ2: Component Analysis.}
What are the respective contributions of the fast and slow monitoring components?

\noindent\textbf{RQ3: Efficiency and Robustness.}
What is the computational overhead of DS-MCM, and how sensitive is it to key hyperparameters?

\noindent\textbf{RQ4: Quality and Generalization.}
Are the corrective suggestions reasonable to humans, and does DS-MCM generalize across different experience memories?

\subsection{Experimental Settings}

\begin{table*}[t]
\centering
\small
\setlength{\tabcolsep}{6pt}
\renewcommand{\arraystretch}{1.15}
\caption{Main results (\%) on deep-search benchmarks. Average is computed across the four benchmarks, with its statistical significance assessed via paired t-tests; missing entries (``--'') are excluded from averaging. Colored cells highlight DS-MCM results, with light orange marking notable cases that surpass proprietary systems. Boldface denotes the best performance. $\dagger$ indicates statistically significant improvements over the corresponding baselines (permutation test with 10{,}000 trials, $p<0.05$).}

\label{tab:main_results}
\resizebox{0.8\linewidth}{!}{
\begin{tabular}{lccccc}
\toprule
\textbf{Model} & \textbf{BrowseComp-Plus} & \textbf{BrowseComp-ZH} & \textbf{X-Bench} & \textbf{GAIA} & \textbf{Average} \\
\midrule
\multicolumn{6}{c}{\textbf{Proprietary Deep-Search Systems / Models}} \\
\midrule
OpenAI GPT-5           & \textbf{70.0} & 34.3 & \textbf{75.0} & 62.8 & \textbf{60.5} \\
OpenAI o3              & 63.5 & 29.1 & 65.0 & \textbf{70.9} & 57.1 \\
Gemini 2.5 Pro         & 59.4 & 27.3 & 56.0 & 60.6 & 50.8 \\
Grok-3 DeeperSearch    & --   & 12.2 & 50.0 & 47.1 & 36.4 \\
OpenAI DeepSearch      & 51.5 & \textbf{42.9} & 66.7 & 70.5 & 57.9 \\
\midrule
\multicolumn{6}{c}{\textbf{Open-Source Deep-Search Systems / Models}} \\
\midrule
Tongyi-DeepResearch            & 51.0 & 38.0 & 69.0 & 70.1 & 57.0 \\
\quad + LLM-Critic             & 55.0 & 42.0 & 68.0 & 69.3 & 58.6 \\
\rowcolor{gray!12}
\quad \textbf{+ DS-MCM}                 
& {\textbf{62.0}$^{\dagger}$}
& \thl{\textbf{45.0}$^{\dagger}$}
& {\textbf{74.0}$^{\dagger}$}
& \thl{\textbf{71.7}}
& \thl{\textbf{63.2}$^{\dagger}$} \\
\midrule
MiroThinker-DeepResearch       & 21.0 & 31.0 & 62.0 & 63.8 & 44.5 \\
\quad + LLM-Critic             & 24.0 & 30.0 & 64.0 & 63.0 & 45.2 \\
\rowcolor{gray!12}
\quad \textbf{+ DS-MCM}                 &
\textbf{26.0} &
\textbf{34.0}$^{\dagger}$ &
\textbf{68.0}$^{\dagger}$ &
\textbf{69.3}$^{\dagger}$ &
\textbf{49.3}$^{\dagger}$ \\
\midrule
Qwen3-30B-MoE                  & 5.0  & 21.0 & 42.0 & 40.9 & 27.2 \\
\quad + LLM-Critic             & 22.0 & 27.0 & 47.0 & 43.3 & 34.8 \\
\rowcolor{gray!12}
\quad \textbf{+ DS-MCM}                 &
\textbf{29.0}$^{\dagger}$ &
\textbf{35.0}$^{\dagger}$ &
\textbf{53.0}$^{\dagger}$ &
\textbf{47.2}$^{\dagger}$ &
\textbf{41.1}$^{\dagger}$ \\
\bottomrule
\end{tabular}}
\end{table*}

\subsubsection{Datasets and Evaluation Metrics}

We evaluate DS-MCM on four deep-search benchmarks: BrowseComp-Plus~\cite{chenbrowsecomp}, BrowseComp-ZH~\cite{zhou2025browsecomp}, xbench-DeepSearch~\cite{chen2025xbench}, and GAIA~\cite{mialon2023gaia}, covering controlled English search, high-difficulty Chinese web search, tool-centric deep search, and general-purpose assistant evaluation.

BrowseComp-Plus~\cite{chenbrowsecomp} replaces live web search with a fixed curated corpus, enabling fair and reproducible assessment of retrieval--reasoning behaviors, while BrowseComp-ZH~\cite{zhou2025browsecomp} extends this setting to the Chinese web with high-difficulty multi-hop questions. xbench-DeepSearch~\cite{chen2025xbench} focuses on tool use and information seeking in complex retrieval-centric scenarios, and GAIA~\cite{mialon2023gaia} evaluates general AI assistants on real-world tasks requiring web browsing, reasoning, and tool use.
Since BrowseComp-Plus and BrowseComp-ZH contain thousands of questions, we follow prior work~\cite{zeng2025pushing} and randomly sample 100 queries from each benchmark for evaluation, providing a cost-efficient yet representative estimate of overall performance.
For the slow monitor (Section~\ref{sec:slow_monitor}), the metacognitive memory set $\mathcal{M}$ is constructed by sampling 500 trajectories from the BrowseComp~\cite{wei2025browsecomp} training split, ensuring no overlap with any evaluation questions.

We adopt \textbf{Accuracy} as the primary evaluation metric following the standard evaluation protocol of each benchmark~\cite{team2025tongyi}.

\subsubsection{Backbone and Baseline Models}
We evaluate DS-MCM on three representative open-source deep-search backbones: \textsc{Tongyi-DeepResearch-A30B-A3B} (\textsc{Tongyi-DeepResearch})~\cite{team2025tongyi}, \textsc{Miro-Thinker-v1.0-8B} (\textsc{MiroThinker-DeepResearch})~\cite{team2025mirothinker}, and \textsc{Qwen3-A30B-A3B-Instruct-2507} (\textsc{Qwen3-30B-MoE})~\cite{yang2025qwen3}. These models cover both full-fledged deep-research agent systems and modern instruction-tuned MoE reasoning backbones, providing a diverse and strong foundation for assessing the generality of our approach.

On each backbone, we implement our proposed DS-MCM and compare it against a strong \emph{LLM-as-Critic} baseline, where a critic model directly performs both fast monitoring and slow reflective analysis through prompted self-evaluation, without explicit uncertainty--evidence consistency modeling or experience-driven metacognitive memory. 

For additional comparison, we also report results from several proprietary deep-search systems and models, including \textsc{GPT-5} and \textsc{o3}~\cite{openai_gpt5_2025} from OpenAI, \textsc{Gemini2.5 Pro}~\cite{google_gemini_deep_research_2025}, \textsc{Grok-3 DeeperSearch}~\cite{xai_grok_2025}, and \textsc{OpenAI Deep Research}~\cite{openai_deep_research_2025}. The reported results for these proprietary systems are obtained from existing benchmarks~\cite{zeng2025pushing,chen2025xbench}.

\subsubsection{Implementation Details}

DS-MCM is implemented on top of a standard ReAct-based deep search agent without modifying the underlying task-level policy. We use \textsc{Qwen-Embedding-8B}~\cite{qwen3embedding} as the unified embedding model for semantic clustering in the Fast Consistency Monitor and for encoding sessions and memory entries in the Slow Experience-Driven Monitor. At each retrieval step, the agent retrieves the top-$5$ documents, which are clustered in the embedding space to compute Searching Entropy. All session and memory embeddings are indexed using a FAISS-based vector index to enable efficient similarity search~\cite{douze2025faiss}.
The Fast Consistency Monitor is applied at every step, with the anomaly threshold set to $\tau = k\sigma$ and $k=2$. When an anomaly is detected, the Slow Experience-Driven Monitor is triggered, using \textsc{Qwen3-A30B-A3B-Instruct-2507} as the Critical Model~\cite{yang2025qwen3} and GPT~5.2~\cite{openai_gpt5_2025} to construct experience memory. For each flagged session, we retrieve two relevant entries from the success memory and two from the failure memory ($|\mathcal{R}_t^{+}|=|\mathcal{R}_t^{-}|=2$) to perform memory-conditioned diagnosis and generate corrective suggestions. 

All experiments are conducted using the \textsc{SGLang} inference engine~\cite{zheng2024sglang} with the HuggingFace \texttt{Transformers} library~\cite{wolf2019huggingface}, on machines equipped with NVIDIA A6000 GPUs and 52-core Intel(R) Xeon(R) Gold 6230R CPUs at 2.10GHz.  In addition to the BrowseComp-Plus benchmark, external documents during deep search are obtained via the Google Search API~\footnote{\url{https://serper.dev/}} and processed using the Jina API~\footnote{\url{https://jina.ai/}}.

\subsection{RQ1: Overall Performance}

Table~\ref{tab:main_results} summarizes the performance of different deep-search systems across four benchmarks. Overall, \textbf{DS-MCM consistently and substantially improves performance across all evaluated backbones}, yielding significant gains over both vanilla deep-search systems and strong LLM-as-Critic baselines. For open-source models, DS-MCM achieves clear improvements on nearly all benchmarks, with many gains being statistically significant (†), demonstrating that explicit meta-cognitive monitoring provides benefits beyond generic LLM as critic alone.

Notably, \textbf{DS-MCM elevates the open-source Tongyi DeepResearch to a new performance regime}. With DS-MCM, Tongyi-DeepResearch attains the highest overall average score among all evaluated systems, surpassing its own LLM-Critic variant by a large margin and \textbf{outperforming multiple proprietary deep-search systems}, including OpenAI~o3, OpenAI~DeepSearch, Gemini~2.5~Pro, and Grok-3~DeeperSearch on average. This result is particularly striking given that these proprietary systems rely on larger closed models and tightly integrated commercial pipelines.

These results indicate that the performance gap between open-source and proprietary deep-search systems is \textbf{not solely determined by model scale or closed infrastructures}. Instead, principled, neuroscience-inspired meta-cognitive monitoring can substantially improve robustness and accuracy, enabling open-source deep-search agents to match or even exceed the performance of state-of-the-art commercial systems.

\subsection{RQ2: Ablation Study of DS-MCM}

\begin{table}[t]
\centering
\caption{Ablation study of DS-MCM on different open-source deep-search backbones.}
\label{tab:ablation}
\resizebox{0.99\linewidth}{!}{
\begin{tabular}{lccc}
\toprule
\textbf{Model} & \textbf{Tongyi-DR} & \textbf{MiroThinker-DR} & \textbf{Qwen3-30B-MoE} \\
\midrule
w/o Experience Memory   & 53 & 23 & 23 \\
w/o Searching Entropy  & 57 & 24 & 25 \\
\rowcolor{gray!12}
\textbf{FULL DS-MCM}   & \textbf{62} & \textbf{26} & \textbf{29} \\
\bottomrule
\end{tabular}}
\end{table}
Table~\ref{tab:ablation} presents an ablation study that disentangles the contributions of the Fast Consistency Monitor and the Slow Experience-Driven Monitor in DS-MCM on BrowseComp-Plus. Removing either component leads to consistent performance degradation across all backbones, confirming that both modules play complementary and non-redundant roles in effective deep-search monitoring.
When the experience memory in the Slow Experience-Driven Monitor is removed (w/o Experience Memory), performance drops substantially on all models. This degradation is particularly pronounced on Tongyi-DeepResearch, indicating that memory-conditioned reflection is critical for diagnosing recurring cognitive patterns and guiding corrective behavior beyond local step-level signals. Without access to accumulated success and failure experiences, the slow monitor degenerates into a context-free critic, limiting its ability to provide targeted and actionable interventions.
In contrast, removing Searching Entropy from the Fast Consistency Monitor (w/o Searching Entropy) also results in noticeable performance declines, though to a lesser extent. This setting reduces fast monitoring to reasoning-entropy-only detection, which fails to account for uncertainty arising from ambiguous or conflicting external evidence. The observed drops demonstrate that calibrating internal reasoning uncertainty against external evidence uncertainty is essential for reliable anomaly detection in deep-search settings, where high reasoning entropy does not necessarily indicate erroneous behavior.


\begin{table}[t]
\centering
\caption{Evaluation on the Who\&When benchmark for identifying faulty agents and erroneous steps.}
\label{tab:whowhen}
\resizebox{\linewidth}{!}{
\begin{tabular}{lcccc}
\toprule
\textbf{Method} &
\multicolumn{2}{c}{\textbf{Who\&When (Handcrafted)}} &
\multicolumn{2}{c}{\textbf{Who\&When (Automated)}} \\
\cmidrule(lr){2-3} \cmidrule(lr){4-5}
 & \textbf{Agent-level} & \textbf{Step-level} & \textbf{Agent-level} & \textbf{Step-level} \\
\midrule
Qwen3-8B                    & 46.55 & 3.45  & 57.14 & 14.29 \\
\rowcolor{gray!12}
\quad + MCM  & \textbf{55.17} & \textbf{22.41} & 57.14 & \textbf{34.13} \\
\midrule
Qwen3-30B                   & 31.03 & 5.17  & 57.94 & 19.05 \\
\rowcolor{gray!12}
\quad + MCM  & \textbf{39.66} & \textbf{15.52} & \textbf{64.29} & \textbf{44.44} \\
\midrule
GPT-4o                      & 51.72 & 5.17  & 41.27 & 14.29 \\
\rowcolor{gray!12}
\quad + MCM  & 51.72 & \textbf{22.41} & \textbf{54.76} & \textbf{38.89} \\
\midrule
GPT-4.1                     & 36.21 & 8.62  & 60.32 & 26.98 \\
\rowcolor{gray!12}
\quad + MCM  & \textbf{55.17} & \textbf{24.14} & 60.32 & \textbf{43.65} \\
\midrule
GPT-5                       & 36.21 & 18.97 & 58.73 & 14.29 \\
\rowcolor{gray!12}
\quad + MCM  & \textbf{51.72} & \textbf{29.31} & \textbf{60.76} & \textbf{50.79} \\
\bottomrule
\end{tabular}}
\end{table}
\subsection{RQ2: Effectiveness of Experience-Driven Monitoring for Agent Process Supervision}
\label{sec:process_localization}

To evaluate the ability of the Slow Experience-Driven Monitor to identify errors during execution, we conduct experiments on the Who\&When benchmark~\cite{zhangagent}, which includes a handcrafted subset derived from Magnetic-One~\cite{fourney2024magentic} and an automated subset constructed from AG2~\cite{ag2ai2024ag2}. Following prior work~\cite{zhangagent}, we report agent-level accuracy, measuring whether the faulty agent is correctly identified, and step-level accuracy, assessing whether the exact erroneous reasoning step is localized. To enforce a strict generalization setting, we adopt a cross-memory protocol: when evaluating on the handcrafted subset, the experience memory is constructed from the automated subset, and vice versa, preventing memory leakage and encouraging reusable error pattern learning.

As shown in Table~\ref{tab:whowhen}, incorporating MCM with experience memory consistently improves error localization performance across all LLM based critical models and both subsets. The most substantial gains occur at the step level, where MCM significantly increases the accuracy of identifying the precise erroneous step, indicating stronger process-level supervision. Agent-level accuracy also improves for most backbones, suggesting better attribution of failures within multi-step trajectories. Notably, these improvements persist under the cross-memory evaluation protocol, demonstrating that MCM captures generalizable execution-level error patterns rather than dataset-specific artifacts. Overall, these results validate the effectiveness of the slow experience-driven monitor in detecting and localizing errors during deep-search execution.

\subsection{RQ3: Sensitivity Analysis}
\begin{figure*}[t]
    \centering

    \begin{subfigure}[t]{0.32\textwidth}
        \centering
        \includegraphics[width=\textwidth, trim=0 8 0 0, clip]{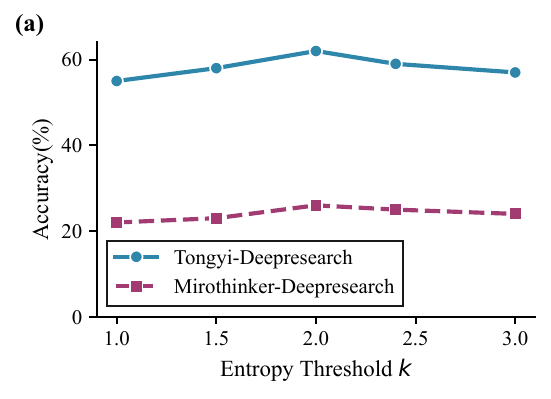}
        \label{fig:sens_entropy}
    \end{subfigure}
    \hfill
    \begin{subfigure}[t]{0.32\textwidth}
        \centering
        \includegraphics[width=\textwidth, trim=0 8 0 0, clip]{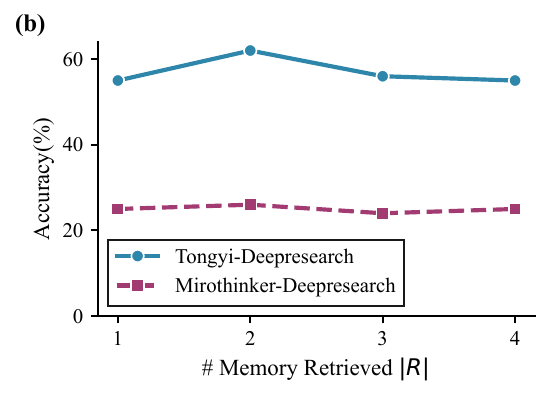}
        \label{fig:sens_memory}
    \end{subfigure}
    \hfill
    \begin{subfigure}[t]{0.32\textwidth}
        \centering
        \includegraphics[width=\textwidth, trim=0 8 0 0, clip]{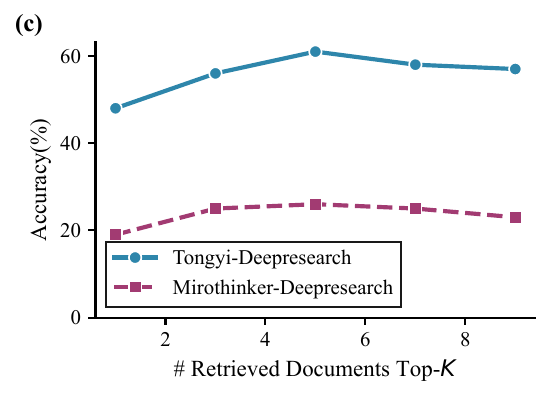}
        \label{fig:sens_docs}
    \end{subfigure}
    \caption{Sensitivity analysis of DS-MCM with respect to key hyperparameters.}
    \label{fig:sensitivity}
\end{figure*}

Figure~\ref{fig:sensitivity} analyzes the sensitivity of DS-MCM to key hyperparameters in both the Fast Consistency Monitor and the Slow Experience-Driven Monitor on BrowseComp-Plus. 
We first vary the entropy threshold $k$ used for anomaly detection in the Fast Consistency Monitor. Performance peaks around $k=2$ for both backbones, while smaller values lead to overly aggressive triggering and larger values delay necessary intervention. This observation validates the use of a moderate threshold that balances sensitivity and robustness.
Next, we examine the number of retrieved memory entries $|R|$ in the Slow Experience-Driven Monitor. Retrieving two memory entries yields the best performance, while further increasing $|R|$ provides diminishing or even negative returns. This suggests that compact and highly relevant experience memory is more effective than larger but noisier retrieval sets.
Finally, we vary the number of retrieved documents Top-$K$ used to compute Searching Entropy. Performance improves as Top-$K$ increases up to a moderate range, reflecting more reliable estimation of evidence uncertainty, but degrades slightly when too many documents are included, likely due to increased semantic noise.
These results indicate that DS-MCM is not overly sensitive to hyperparameter choices and that its performance gains arise from principled monitoring rather than fine-tuned parameter selection.

\subsection{RQ3: Time Efficiency Analysis}
\begin{table}[t]
\centering
\caption{Time efficiency analysis (seconds per query) on deep-search benchmarks.
Numbers in parentheses indicate relative overhead compared to the corresponding backbone.}
\label{tab:efficiency}
\resizebox{0.99\linewidth}{!}{
\begin{tabular}{lcc}
\toprule
\textbf{Method} & \textbf{BrowseComp-Plus} & \textbf{GAIA} \\
\midrule
Tongyi-DeepResearch            & 519.71 (×1.00) & 219.57 (×1.00) \\
\quad + LLM-Critic             & 581.43 (×1.12) & 253.43 (×1.15) \\
\rowcolor{gray!12}
\quad + DS-MCM                 & 541.43 (×1.04) & 235.35 (×1.07) \\
\midrule
MiroThinker-DeepResearch       & 608.76 (×1.00) & 273.56 (×1.00) \\
\quad + LLM-Critic             & 690.32 (×1.13) & 334.23 (×1.22) \\
\rowcolor{gray!12}
\quad + DS-MCM                 & 631.25 (×1.03) & 292.41 (×1.07) \\
\bottomrule
\end{tabular}
}
\end{table}

Table~\ref{tab:efficiency} reports the runtime overhead of different monitoring strategies on BrowseComp-Plus and GAIA. Incorporating an LLM-Critic consistently introduces substantial latency, increasing end-to-end execution time by 12--22\% across backbones and benchmarks. This overhead arises from invoking a full critic model at every step, regardless of whether reflective intervention is necessary.
In contrast, DS-MCM incurs only modest additional cost. Across both Tongyi-DeepResearch and MiroThinker-DeepResearch, DS-MCM increases runtime by merely 3--7\%, which is significantly lower than that of LLM-Critic. This efficiency gain stems from the hierarchical design of DS-MCM: the Fast Consistency Monitor operates with lightweight detection at every step, while the computationally expensive Slow Experience-Driven Monitor is triggered only when abnormal execution states are detected.
These results demonstrate that DS-MCM achieves a favorable trade-off between effectiveness and efficiency. By avoiding unconditional critique and enabling selective reflective control, DS-MCM delivers strong performance improvements with minimal additional latency, making it more suitable for practical deployment of deep-search agents.

\begin{table}[t]
\centering
\caption{Effect of different knowledge bases for constructing the Slow Experience-Driven Monitor.
All results are evaluated on BrowseComp-Plus. $\Delta$ denotes the absolute improvement over the corresponding base model.}
\label{tab:memory_source}
\resizebox{0.9\linewidth}{!}{
\begin{tabular}{lccc}
\toprule
\textbf{Method} & \textbf{Knowledge Base} & \textbf{Accuracy (\%)} & $\boldsymbol{\Delta}$ \\
\midrule
\multicolumn{4}{l}{\textit{Tongyi-DeepResearch}} \\
Base & -- & 51 & -- \\
\rowcolor{gray!6}
\quad + DS-MCM & BrowseComp & 62 & 0.11 \\
\rowcolor{gray!12}
\quad + DS-MCM & GAIA       & 60 & 0.09 \\
\midrule
\multicolumn{4}{l}{\textit{MiroThinker-DeepResearch}} \\
Base & -- & 21 & -- \\
\rowcolor{gray!6}
\quad + DS-MCM & BrowseComp & 26 & 0.05 \\
\rowcolor{gray!12}
\quad + DS-MCM & GAIA       & 25 & 0.04 \\
\bottomrule
\end{tabular}}
\end{table}
\subsection{RQ4: Generalization of Experience Memory}
We further investigate the impact of using different datasets to construct the experience memory for the Slow Experience-Driven Monitor. While our main experiments build the memory using BrowseComp trajectories, we additionally construct an alternative memory using GAIA and evaluate both settings on BrowseComp-Plus.
As shown in Table~\ref{tab:memory_source}, DS-MCM consistently improves over the base models regardless of the memory source. Although using BrowseComp yields slightly stronger gains, memory constructed from GAIA still provides substantial improvements across both backbones. These results indicate that the proposed slow monitor does not rely on dataset-specific artifacts, but instead captures reusable cognitive patterns that transfer across different benchmarks. This flexibility allows DS-MCM to adapt to diverse memory repositories, supporting its applicability in practical and evolving deep-search scenarios.

\subsection{RQ4: Human Evaluation}
\begin{table}[t]
\centering
\caption{Human evaluation of corrective suggestions $\delta_t$ generated by different monitoring strategies.}
\label{tab:human_eval}

\begin{tabular}{lcc}
\toprule
\textbf{Method} & \textbf{BrowseComp-Plus} & \textbf{GAIA} \\
\midrule
\multicolumn{3}{l}{\textit{Tongyi-DeepResearch}} \\
\quad + LLM-Critic & 69.0 & 65.0 \\
\rowcolor{gray!12}
\quad + DS-MCM     & \textbf{80.0} & \textbf{73.0} \\
\midrule
\multicolumn{3}{l}{\textit{MiroThinker-DeepResearch}} \\
\quad + LLM-Critic & 64.0 & 59.0 \\
\rowcolor{gray!12}
\quad + DS-MCM     & \textbf{75.0} & \textbf{79.0}  \\
\bottomrule
\end{tabular}
\end{table}

We conduct a human evaluation to assess whether the corrective suggestions $\delta_t$ generated by the Slow Experience-Driven Monitor are reasonable. We randomly sample 100 sessions where the slow monitor is triggered and provide annotators with the query, execution history, and current session state. Each session is independently evaluated by two human annotators, and a suggestion is counted as correct only when both annotators agree.
As shown in Table~\ref{tab:human_eval}, DS-MCM consistently outperforms the LLM-Critic baseline across both backbones and benchmarks.  These results indicate that the experience-driven slow monitor produces more reasonable and actionable corrective suggestions than generic LLM-based critique.

\section{Conclusion}
Deep-search agents operate under long-horizon execution with continuously evolving and often uncertain external evidence, making effective monitoring and regulation critical for reliable performance. However, existing systems largely focus on task-level reasoning while lacking explicit mechanisms for execution-level metacognitive control.
We present DS-MCM, a metacognitive monitoring framework for deep-search agents that separates task-level reasoning from execution-level control. DS-MCM integrates a fast consistency check between reasoning and evidence uncertainty with a slow, experience-driven monitor to guide corrective behavior. Experiments show that DS-MCM consistently improves performance across benchmarks and backbones, enabling open-source systems to rival or surpass proprietary deep-search models. Further analyses confirm the complementary roles of fast and slow monitoring with only modest computational overhead. 
These results highlight metacognitive monitoring as a key ingredient for robust and practical deep-search agents.

\bibliographystyle{ACM-Reference-Format}
\bibliography{sample-base}

\appendix

\end{document}